\documentclass{article}

\usepackage{arxiv}

\usepackage[utf8]{inputenc} % allow utf-8 input
\usepackage[T1]{fontenc}    % use 8-bit T1 fonts
\usepackage{hyperref}       % hyperlinks
\usepackage{url}            % simple URL typesetting
\usepackage{booktabs}       % professional-quality tables
\usepackage{amsfonts}       % blackboard math symbols
\usepackage{nicefrac}       % compact symbols for 1/2, etc.
\usepackage{microtype}      % microtypography
\usepackage{lipsum}
\usepackage{caption}
\usepackage{subcaption}
\usepackage{graphicx}
\usepackage{tcolorbox}
\usepackage[normalem]{ulem}

\title{\mbox{Problem Learning: Towards the Free Will of Machines}}
\author{
Yongfeng Zhang \\
Department of Computer Science\\
Rutgers University, New Brunswick, NJ 08854 \\
\texttt{yongfeng.zhang@rutgers.edu}
}
\begin{document}

\maketitle

\begin{abstract}
A machine intelligence pipeline usually consists of six components: problem, representation, model, loss, optimizer and metric. 
% Take sentiment (or image) classification as an example, the \textit{problem} is to classify sentences into various sentiment labels. To solve the problem, we convert words or sentences into \textit{representations} such as vectors. The representations will be processed by a \textit{model} such as LSTM and fed into a \textit{loss function} such as cross-entropy loss, which characterizes the quality of our current representation and model. Then, an \textit{optimizer} such as back-propagation is used to optimize the loss for the best parameters. The representation and model can sometimes be integrated in a unified architecture, and the representations can be either manually designed (e.g., TF-IDF) or automatically learned (e.g., word embeddings). Finally, we use some (usually manually designed) \textit{evaluation metrics} such as accuracy, precision, recall and AUC to evaluate the task.
Researchers have worked hard trying to automate many components of the pipeline. For example, Representation Learning aims to automatically learn good features from data, Model Learning (or Neural Architecture Search) aims to find good model architectures for a task, Loss Learning aims to discover good loss functions, Learning to Optimize automatically discovers good optimization algorithms, and Learning to Evaluate discovers learning-based evaluation metrics in place of manually designed rule-based metrics.

However, one key component of the pipeline---problem definition---is still left mostly unexplored in terms of automation. 
Usually, it requires extensive efforts from domain experts to identify, define and formulate important problems in an area.
% such as classification, prediction, ranking, etc. in a specific area, not only for traditional AI  areas such as vision and language processing, search and recommendation systems, etc., but also for many emerging research areas such as smart city, smart health and intelligent economic analysis. 
This is partly because the community has yet to solve many existing manually defined problems rather than automatically detecting new problems. However, automatically discovering research or application problems for an area is beneficial since it helps to identify valid and potentially important problems hidden in data that are unknown to domain experts, expand the scope of tasks that we can do in an area, and even inspire completely new findings.

This paper describes Problem Learning, which aims at learning to discover and define valid and ethical problems from data or from the machine's interaction with the environment. We formalize problem learning as the identification of valid and ethical problems in a problem space and introduce several possible approaches to problem learning, such as problem acquisition, problem search, problem generalization, problem composition and problem decomposition. 
Besides, problem definition is usually closely related with evaluation, because we need to know how successful a potential solution to the problem is. As a result, we take learning to evaluate to accompany problem learning so as to automate the problem definition and evaluation pipeline.

In a broader sense, problem learning is an approach towards the free will of intelligent machines. Currently, machines are still limited to solving the problems defined by humans, without the ability or flexibility to freely explore various possible problems that are even unknown to humans. Though many machine learning techniques such as representation learning and neural architecture search have been developed and integrated into intelligent systems, they still focus on the means rather than the purpose in that machines are still solving human defined problems. However, proposing good problems is sometimes even more important than solving problems, because a good problem can help to inspire new ideas and gain deeper understandings. The paper also discusses the ethical implications of problem learning under the background of Responsible AI.

% The conquest of machine intelligence needs to automate various components in the pipeline, five components: problem, model, representation, loss, optimizer, metric (evaluation). 
% Representation Learning, Model Learning, Loss Learning have been pretty much automated.
% However, the problem definition is still pretty much manually, request domain experts to identify, define and formulate important problems in each domain. 
% Problem learning aims to .
% Benefits: 1. automate, discover potential new and important problems. 
% Free will of machines.
% In this paper, we ...

% A complete machine learning pipeline works in this way: . Most of ... has been automated, some are being automated. The only missing component is problem definition. 

% Usually, domain experts manually identify important problems, carefully define the problem, formulate the problem, and define how a potential solution to the problem should be evaluated. On one hand, one the other hand, machines are limited in the sense that they are limited to solve problems defined by humans. 

% Problem definition is usually closely bonded with how to evaluate the problem. Problem learning + learning to evaluate. 
\end{abstract}

% keywords can be removed
\keywords{Problem Learning \and Representation Learning \and Model Learning \and Loss Learning \and Learning to Evaluate \and AI}

\section{Introduction}
\label{sec:introduction}
% Problem learning aims at learning to define a problem.

A machine intelligence pipeline usually consists of six components: problem, representation, model, loss, optimizer and metric. Take sentiment or image classification as an example \cite{deng2009imagenet,krizhevsky2012imagenet,pang2008opinion,liu2012sentiment}, the \textit{problem} is to classify sentences or images into various sentiment or image class labels. To solve the problem, we convert words, sentences or images into \textit{representations} such as vectors \cite{mikolov2013distributed}. The representations will be processed by a \textit{model} such as Long Short-Term Memory (LSTM) \cite{hochreiter1997long} or Convolutional Neural Network (CNN) \cite{kalchbrenner2014convolutional,lecun1995convolutional} and fed into a \textit{loss function} such as cross-entropy loss \cite{murphy2012machine}, which characterizes the quality of the current representation and model. Then, an \textit{optimizer} such as back-propagation \cite{rumelhart1986learning} or Stochastic Gradient Descent (SGD) \cite{bottou2010large} is used to optimize the loss for the best parameters. The representation and model can sometimes be integrated in a unified architecture \cite{lecun2015deep}, and the representations can be either manually designed (e.g., TF-IDF \cite{robertson2004understanding}) or automatically learned (e.g., word embeddings \cite{mikolov2013distributed}). Finally, we can use some (usually manually designed) \textit{evaluation metrics} such as accuracy, precision, recall and F-measure to evaluate the task \cite{powers2011evaluation}.

% This paper discusses possible approaches, benefits, consequences and ethical considerations of problem learning. 

Researchers have worked hard to automate many components of the pipeline. For example, Representation Learning \cite{hinton1986learning,bengio2013representation,hamilton2017representation,scholkopf2021toward} aims to automatically learn good features from data, Model Learning (Neural Architecture Search, or broadly speaking Automatic Machine Learning) \cite{zoph2017neural,pham2018efficient,liu2019darts} aims to learn good model architectures for a task, Loss Learning \cite{gonzalez2020improved,li2021auto,liu2021loss} aims to learn good loss functions, Learning to Optimize \cite{andrychowicz2016learning,li2017learning,ravi2017optimization} aims to automatically discover good optimization algorithms, and Learning to Evaluate \cite{lowe2017towards,cui2018learning} aims to discover learning-based evaluation metrics in place of manually designed rule-based metrics.

However, one key component of the pipeline---problem definition---is still left mostly unexplored in terms of automation. 
In the current AI research paradigm, it usually requires extensive efforts from domain experts to identify, define and formulate the important problems of a research or application area, and the problem is usually formalized into one of the standard formats such as classification, regression, generation, prediction, ranking, etc. Notable examples include image classification in vision research \cite{deng2009imagenet,krizhevsky2012imagenet}, sentiment classification in natural language research \cite{pang2008opinion,liu2012sentiment}, entity ranking and link prediction in knowledge graph and graph neural networks \cite{bordes2013translating,kipf2017semi,hamilton2017inductive,velivckovic2018graph}, as well as document ranking \cite{salton1975vector,robertson2009probabilistic,salton2003information,ponte1998language} and item ranking \cite{resnick1994grouplens,sarwar2001item,koren2009matrix,cheng2016wide,chen2021neural} in information retrieval and recommender systems. Meanwhile, some relatively complicated problems are usually manually decomposed into several steps of relatively simple problems, e.g., sentence generation is sometimes represented as multiple steps of word ranking problems through beam search \cite{bahdanau2015neural}, and link prediction in knowledge graph is sometimes decomposed into and evaluated as entity ranking problems \cite{bordes2013translating}.
% and many planning tasks are represented as a sequence of smaller actions such as prediction or classification.

Such significant requirement for manual efforts in terms of identifying, defining and formulating problems not only exists for ``traditional'' AI areas such as vision and language processing, search and recommendation systems, etc., but also for many emerging areas such as smart city, smart health and smart economy. This may be partly because the community has yet to solve many existing manually defined problems rather than automatically detecting new problems. However, automatically discovering research or application problems from data is beneficial since it helps to identify valid and potentially important problems hidden in data that are unknown to domain experts, expand the scope of tasks that we can do in an area, and even inspire completely new findings. This is especially important for various emerging AI areas because compared to traditional AI areas where the research problems are usually formalized as ``standard'' tasks, the emerging areas may present a lot more yet unknown problems and sometimes it could be difficult to identity these problems, let alone formalize the problems as one of the standard tasks.

This paper describes Problem Learning, which aims at learning to discover and define valid and ethical problems from data or from the machine's interaction with the environment. We formalize problem learning as the identification of valid and ethical problems in a problem space and we introduce several possible approaches to problem learning such as problem learning from failure, problem learning from exploration, problem composition, problem generalization, problem architecture search and meta problem learning. Besides, problem definition is usually closely related with problem evaluation, because we need to know how successful a potential solution to the problem is. As a result, we take learning to evaluate to accompany problem learning so as to automate the problem definition and evaluation pipeline.

% how to evaluate the problem. Problem learning + learning to evaluate. 

% However, detecting new problems is important ... However, it has been possible to develop problem learning methods for automatically detecting new problems. 

In a broader sense, problem learning is an approach towards the free will of intelligent machines. Currently, machines are still limited to solving the problems defined by humans, without the ability or flexibility to freely explore various possible problems that are even unknown to humans. Though many machine learning techniques such as representation learning and neural architecture search have been developed and integrated into intelligent systems, they still focus more on the means rather than the purpose in the sense that machines are still solving human defined problems. However, proposing good problems is sometimes even more important than solving problems, because a good problem can help to inspire new ideas and gain deeper understandings \cite{newell1972human,simon2012models,schank2013scripts,hutchins1995cognition,pinker2021mind}. Throughout history, the advancement of our science and technology has been constantly driven by new and insightful problems that lead to the development of innovative theories and techniques. As an even more ambitious vision, intelligent machines will quite possibly be set to explore places where human-beings are unable to physically reach such as deep space, deep ocean, and deep subsurface. In this process machines may encounter problems that humans have never imaged. As a result, the ability to identify, define and formalize such new problems through problem learning is a necessity for intelligent machines to survive in such new, unknown environments.

% Explore the universe.

% and Humans create problems/needs and solve the problems. 

Another equally important note to make is the ethical considerations in problem learning. Granting machines with the free will to define problems that they think are important does not mean AI should have the freedom to define and solve \textit{any} problem, but \textit{ethical} problems. This means that problem learning should be conducted under ethical principles \cite{jobin2019global,taddeo2018ai} such as transparency, privacy, fairness, justice, trustworthiness, responsibility and sustainability. For example, some problems such as trying to predict one's sexual orientation, political stance or whether one will quit the job
% or life expectancy
may need to be eliminated from the problem space due to ethical concerns, while some other problems such as predicting 
% whether one will quit the job or 
one's life expectancy may remain in the problem space since they could be helpful to human but they need to be handled with extreme care \cite{siegel2020when}.
% of ethical problems, and some other problems such as predicting 
In general, AI should seek for problems that are non-maleficence, socially beneficial and that advocate human dignity.
% For example, machines should not define problems that. transparency and explainability.
% Sometimes, may not be be a problem. 
% We also discuss the ethical implications of problem learning under the background of Responsible AI.
% The following part of the paper will describe problem learning, approaches to problem learning and ethical considerations of problem learning in more detail.

% \section{From Problem Solving to Problem Creating}

% One could be surprised with the calling for problem creating with AI. For years, we have been developing AI methods that solve or assist in solving problems rather than creating problems. Why do we want to develop intelligent systems that seek to define new problems to solve beyond merely solving manually defined problems?

% The problem creating machine: Image a case, current AI prototype problems. However, while . recommender system example?

% Take recommender system as an example, 

% Find new problems, e.g., detecting what is predictable in data. Note this is different from really designing prediction model, but just detect whether some feature is predictable or worth predicting at all. 

% It has been technically possible to do problem creating, e.g., learning to define problems. 

% \section{Why Problem Learning}
% 1. Help us to discover new, interesting and important problems. e.g., unknown problems in data. Unknown bias in data.

% 2. Free will of machines.

\begin{figure*}[t]
  \centering
  \begin{subfigure}[t]{0.4\textwidth}
        \centering
        \includegraphics[scale=0.5]{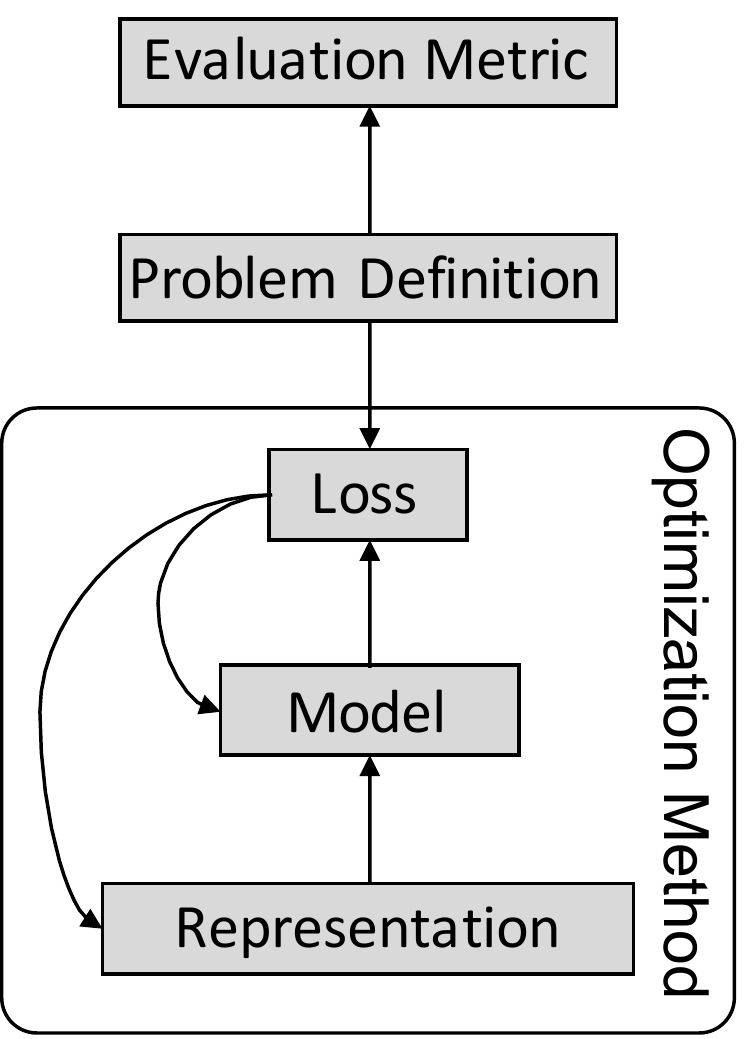}
        \caption{A typical architecture of model AI systems. We first manually define a problem, and the problem decides the corresponding evaluation metric and loss function. Representations and a model are used to calculate the loss, which is minimized by some optimization algorithm so as to optimize the model and representations through back-propagation.}
        \label{fig:architecture1}
    \end{subfigure}
    \hspace{10pt}
    \begin{subfigure}[t]{0.4\textwidth}
        \centering
        \includegraphics[scale=0.5]{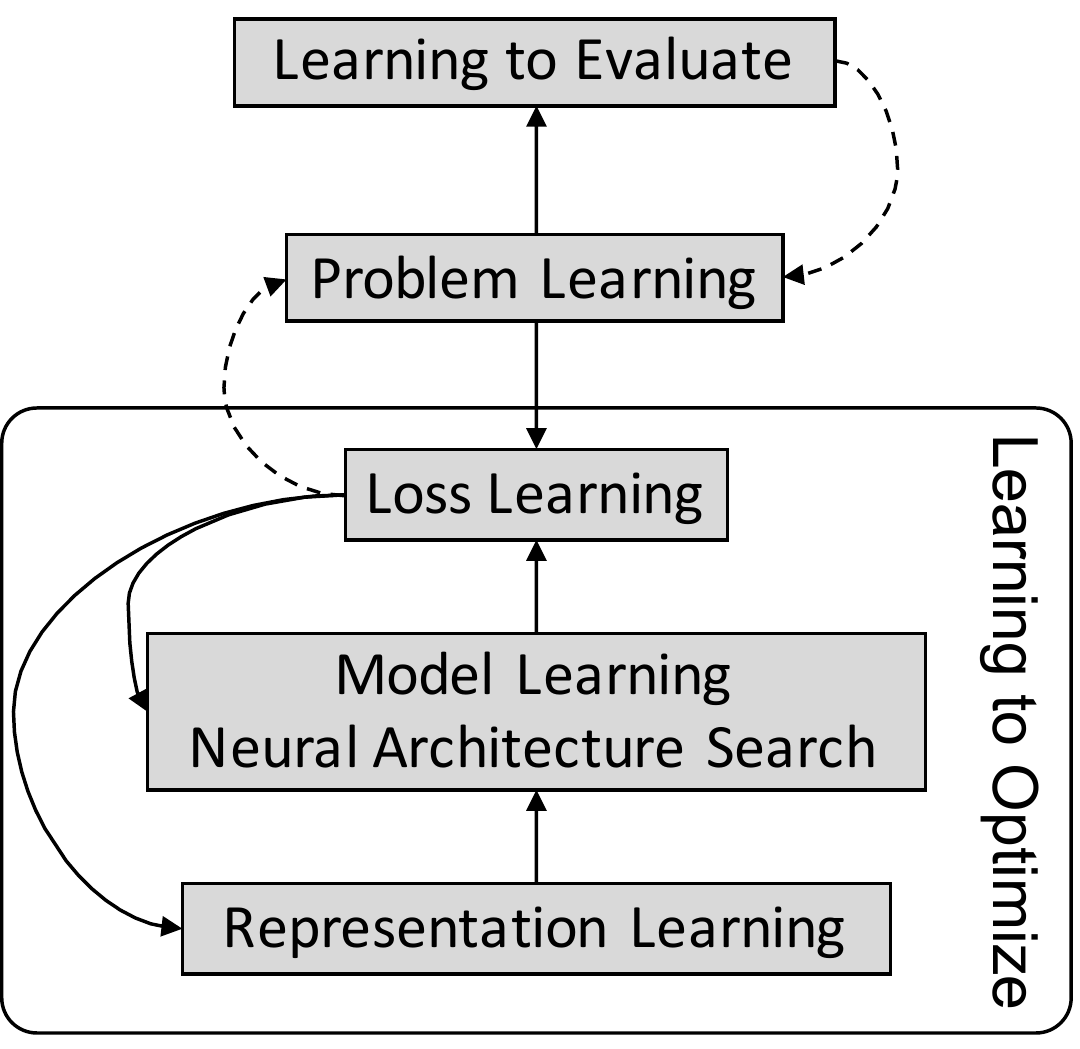}
        \caption{Automate the AI pipeline by incorporating learning into each component of the architecture. Machine decides the problem to solve through learning to define problems, and takes loss/model/representation learning to decide the loss, model and representations needed to solve the problem. It finally evaluates the problem through learning to evaluate.}
        \label{fig:architecture2}
    \end{subfigure}
    \caption{A typical architecture of model AI and automate the pipeline by incorporating learning into each component.}
    \label{fig:architecture}
\end{figure*}

\section{A Hierarchical AI Architecture}
% {of the Machine Learning Framework}

From an abstract point of view, modern AI systems can be represented by a hierarchical architecture, as shown in Figure \ref{fig:architecture1}. In this paradigm, researchers or practitioners first define a problem of interest. For example, the problem can be classifying a bunch of images, sentences or graphs into several class labels, 
% translating a bunch of sentences into another language,
predicting the future values of a time series,
or ranking the potential answers for a given question. Once the problem is clearly defined, researchers are usually aware of how to evaluate a potential solution to the problem. For example, classification problems can be evaluated using some widely used measures such as precision, recall, and accuracy \cite{powers2011evaluation}, prediction problems can be evaluated using some prediction error metrics such as MSE, RMSE or MAE \cite{willmott2005advantages}, and ranking problems can be evaluated using many ranking metrics such as NDCG, MRR and Hit-Ratio \cite{jarvelin2002cumulated}. Sometimes the model can also be evaluated based on human judgements. Researchers then design a loss function that reflects the nature of the problem as much as possible. For example, classification problems usually can be formulated into a cross-entropy loss \cite{murphy2012machine}, prediction problem can be formulated into a prediction error loss \cite{harvey1990forecasting}, and ranking problems can be formulated into a pair-wise ranking loss \cite{rendle2012bpr} or contrastive loss \cite{chen2020simple}. Sometimes, a combination of several loss functions or some adjusted forms of the loss functions are used to model complicated problems. The input to the loss function is produced by some model that is carefully designed for the problem, such as convolutional neural networks \cite{kalchbrenner2014convolutional,lecun1995convolutional}, recurrent neural networks \cite{hopfield1982neural,rumelhart1986learning,hochreiter1997long,goodfellow2016deep}, graph neural networks \cite{kipf2017semi,hamilton2017inductive,velivckovic2018graph}, Transformers \cite{vaswani2017attention,devlin2019bert}, etc., and the model operates over the representations, which are usually distributed vector embeddings of the raw input.
% such as vector embeddings.

% Starting from the fundamental representation part, 
The community has been putting efforts to automate many components of the architecture, as shown in Figure \ref{fig:architecture2}. A long-term vision on AI is to automate each and every component of the pipeline so that machines can automatically identify a problem (problem learning), automatically build solutions to the problem (representation learning, model learning, loss learning, learning to optimize), and finally automatically evaluate the solutions (learning to evaluate).
% making way towards the free will of machines. 
Problem learning, allowing machines to automatically identify the problems that they think are important and worth solving, is the last missing piece of the puzzle and the key component towards the free will of machines. Meanwhile, it is also the component of the pipeline that significantly requires careful considerations in terms of ethical and responsible AI so that a machine with free will helps rather than hurts.

\textbf{Representation Learning}: Early intelligent systems mostly used manually designed representations. A typical example is the TF-IDF vector representation (or bag-of-words representation) for text, which is widely used in many areas such as Information Retrieval, Natural Language Processing, Web Data Mining, among others \cite{robertson2004understanding}. Similarly, in Computer Vision research, extracting hand-crafted features from image was a major step for image processing in the pre-deep learning era \cite{o2019deep}. The extensive efforts needed in feature engineering drives researchers to think about whether it would be possible to let machines learn the features from data automatically. Representation learning \cite{bengio2013representation} serves this goal. Through end-to-end learning, a deep model is ``trained'' on the given data, which discovers the underlying patterns in the training data and automatically extracts the most descriptive and salient features, usually in the form of representation vectors \cite{o2019deep}. Sometime, the extracted features can be organized in a hierarchy to represent different levels of abstraction \cite{bengio2013representation,scholkopf2021toward}. Representation learning greatly alleviated the manual efforts in the most fundamental layer of the hierarchical AI architecture (Figure \ref{fig:architecture2}).

\textbf{Model Learning}: Automated model designing is the next step of automating the AI pipeline. For years researchers have spent countless of efforts to design good models for various tasks. Most of the models, no matter simple or complicated, are designed with handcrafted structures based on the researcher's insightful understanding of the problem at hand. Early models tend to use shallow structures, for example, Linear Regression \cite{montgomery2021introduction} and linear Support Vector Machine \cite{cortes1995support} are structured as a weighted linear summations over the input variables. More recent models usually rely on deep neural networks to build the model structure. For example, Convolutional Neural Networks use multiple layers of convolution filters to build the neural structure and capture the local connectivity \cite{kalchbrenner2014convolutional,lecun1995convolutional}, Recurrent Neural Networks repeat a neural structure over the input sequence to capture the sequential connectivity \cite{hopfield1982neural,rumelhart1986learning,hochreiter1997long,goodfellow2016deep}, Graph Neural Networks adopt several layers of message passing and aggregation to build the neural structure and capture the connectivity between nodes \cite{kipf2017semi,hamilton2017inductive,velivckovic2018graph}, and Transformer-based language models \cite{vaswani2017attention,devlin2019bert} adopt multiple layers of self-attention and feed-forward structures to capture the relationships among the words in a sentence. Though the model parameters are learned, the model structures are manually designed, which requires significant human efforts. Researchers have been exploring whether it is possible to make the machines automatically learn the optimal model structure for a given task. Neural Architecture Search (NAS) works towards this goal. Some research have shown that automatically assembled model structures can be comparable with or even
better than expert manually designed structures on many tasks \cite{zoph2017neural,pham2018efficient,liu2019darts,liu2018progressive}, alleviating the human efforts in terms of model designing.

\textbf{Loss Learning}: Loss function decides how the model output is penalized and provides signals to optimize the model or representation parameters, usually through back propagation. Most of the loss functions are meticulously designed according to domain experts' understanding of the task. As stated above, some ``standard'' tasks are usually mapped to ``standard'' losses, such as prediction to square error loss \cite{harvey1990forecasting}, classification to cross-entropy loss \cite{murphy2012machine}, and ranking to pair-wise ranking loss or contrastive loss \cite{rendle2012bpr}. In many intelligent systems such as
% For single-task loss functions, for examples, mean square error loss is frequently used for prediction tasks, cross-entropy loss is frequently used for . In other applications such as 
smart city and smart health, designing the appropriate loss function is one of the key steps because domain experts' understanding of the problem and their system designation goals are usually reflected by the loss function. Sometimes the loss function can be a combination of several losses in the form of $L=\sum_i\lambda_iL_i$ so as to jointly consider multiple tasks. 
% , the loss function usually reflects what the domain expert's want out of the problem. 
Meanwhile, researchers have found that even for the same task, different losses could have very different performances. For example, recent research found that contrastive loss can be better than cross-entropy loss on many tasks \cite{chen2020simple,you2020graph}. This makes researchers wonder whether it would be possible to make machines automatically learn the optimal loss function for a task. Loss Learning thus aims at this goal. By automatically assembling the loss function from basic operators to find good or even better loss functions compared with manually designed ones \cite{gonzalez2020improved,li2021auto,liu2021loss}, loss learning alleviates the human efforts in terms of loss designing.
% e.g., ..., as a result, we wonder whether we can automatically learn the most optimal loss function for a task. 
% muti-task, add up loss functions. 

\textbf{Learning to Optimize}: Optimization algorithm is the key to learning the model and representation parameters. Many optimization algorithms are meticulously designed by experts and usually based on gradient descent or its variants \cite{lecun1998gradient}. Some examples include stochastic gradient descent \cite{bottou2010large}, momentum methods \cite{tseng1998incremental}, Rprop \cite{riedmiller1993direct}, Adagrad \cite{duchi2011adaptive}, and ADAM \cite{kingma2015adam}. However, algorithm design is a laborious process and often requires many iterations of ideation and validation \cite{li2017learning}. As a result, researchers have been trying to make machines automatically learn good optimization algorithms, for example, using learned rather than designed gradients for parameter updating based on LSTM \cite{andrychowicz2016learning} or reinforcement learning \cite{li2017learning}, which have been shown to be more efficient, accurate and robust than many manually designed optimization algorithms. 

\textbf{Learning to Evaluate}: Once a solution is produced for a given problem, it is important to evaluate the solution so as to know its quality and usability. Many evaluation methods are manually designed rule-based metrics. Except for those commonly known metrics such as accuracy, precision, recall, F-measure, NDCG, MRR, etc. \cite{powers2011evaluation}, researchers sometimes also design tailored metrics for specific tasks. For example, to evaluate image captioning, the semantic propositional image caption evaluation (SPICE) metric \cite{anderson2016spice} is designed to measure the similarity between scene graphs and reference sentences, which shows better correlation with human judgments; to evaluate the explanations of personalized recommendations, the unique sentence ratio (USR) and feature converge ratio (FCR) metrics \cite{li2020generate,li2021personalized} are designed, which are better than generic sentence evaluation metrics such as BLEU and ROUGE on the explainable recommendation \cite{zhang2020explainable,zhang2014explicit} task. However, manually designing the evaluation method is time consuming and the designed metrics are sometimes difficult to generalize to other tasks. Learning to evaluate aims to solve the problem by making machines automatically design the evaluation protocol for a task. 
% automatically learning how to evaluate AI tasks. 
For example, the automatic dialogue evaluation model (ADEM) learns an automatic evaluation procedure for dialogue research \cite{lowe2017towards}, while \cite{cui2018learning} proposed a learning-based discriminative evaluation metric that is directly trained to distinguish between human and machine-generated captions of images. Besides, recent advances on causal machine learning has made it possible for learning to evaluate AI systems based on counterfactual reasoning \cite{ge2021counterfactual}. Some simulation-based evaluation approaches also helps learning to evaluate, which builds a simulation platform to evaluate the intelligent agents such as robotics \cite{choi2021use} and recommender systems \cite{ie2019recsim} in a simulated environment. Learning to evaluate helps to alleviate the manual efforts in designing evaluation protocols or collecting evaluation samples.

\textbf{Problem Learning}: Problem learning is the last missing piece of the puzzle towards an automated AI pipeline, which aims to actively discover and define what problems to solve.
% from data or from the machine’s interaction with the environment.
Problem learning is unique in the sense that it is the key component towards the free will of machines. The other components in the pipeline mostly focus on how to solve a given problem but less on what problem to solve, since the problem is still identified and defined by human, especially by domain experts. However, problem learning drives the behavior of intelligent machines by giving machines the ability and flexibility of deciding what problems they want to solve---a major step towards subjective consciousness. We will discuss problem learning in the following sections.

% and supervises the intelligent machine by telling the agent what problems to solve ---
%Automatic Theorem Discovery. change to learning paradigm.

\section{Problem Learning}

% \subsection{A Formalization of Problems: How to Represent a Problem}
% \subsection{Formalization: How to Represent a Problem}

To provide a formal definition of problem learning, we present three progressive concepts: Solution (S), Problem (P), and Problem Learning (PL).

\subsection{Solution}

% A problem is to do a certain thing on a certain data.

For different AI tasks, solutions may present very different forms, but abstractly, a solution can be usually represented as a mapping that maps questions to answers.

\begin{tcolorbox}[title = Solution (S)]
A solution $S$ is a mapping from the question set to the answer set: $S: \mathbb{Q}\rightarrow \mathbb{A}$
\end{tcolorbox}

To better understand the concept, we use sentiment classification as a example, where the question set $\mathbb{Q}$=\{\textit{All sentences under consideration}\} and the answer set $\mathbb{A}$=\{\textit{Positive (+), Neutral (0), Negative (-)}\}. We can develop various methods for sentiment classification and the final solution would be a mapping from $\mathbb{Q}$ to $\mathbb{A}$, which assigns a sentiment label to each sentence in $\mathbb{Q}$. Sometimes we may adapt the $\mathbb{Q}$ and $\mathbb{A}$ sets to account for specific considerations, e.g., we may add an \textit{Unknown} label to $\mathbb{A}$ so as to account for cases where a sentence in $\mathbb{Q}$ cannot be assigned any sentiment label by the model. Actually, under many machine learning contexts, a solution can be simplified as a mapping from the set of data samples to the set of labels $S: \mathbb{D}\rightarrow \mathbb{L}$. However, we use $S: \mathbb{Q}\rightarrow \mathbb{A}$ for generality because in some cases the questions in $\mathbb{Q}$ may not be simplified as data samples and the answers in $\mathbb{A}$ may not be simplified as labels, which we will discuss in the following sections. 
% \yz{TODO}

An important note to make is that the use of mapping as a mathematical formalization for solutions implies an important assumption, i.e., each element in the question set $\mathbb{Q}$ is mapped to one and only one element in the answer set $\mathbb{A}$ per the mathematical definition of mapping. However, some problems, from the first impression, may require solutions of one-to-many mappings, which violates the definition of mapping. For example, many search or recommendation tasks require a ranking list as the solution, which is an (ordered) set of elements. Consider search engine as an example, $\mathbb{Q}$ is the set of all possible queries and $\mathbb{A}$ is the set of all possible documents. A search result for query $q\in\mathbb{Q}$ would be a subset of documents $\{d\}\subseteq\mathbb{A}$ that are relevant to the query $q$. However, mapping an element in $\mathbb{Q}$ to many elements in $\mathbb{A}$ is prohibited by the definition of mappings.

One possible method to solve the problem is to use set-valued mapping, i.e., we can define $S$ as $\mathbb{Q}\rightarrow 2^{\mathbb{A}}$, where $2^{\mathbb{A}}$ is the power set of $\mathbb{A}$. In this way, a solution $S$ is still a mapping which maps a query to a set of documents. However, implementing this solution is too expensive due to the huge size of the power set $2^{\mathbb{A}}$. A better way is to define the question set as all of the query-document pairs $\mathbb{Q}=\{(q,d)\}$ and define the answer set as the potential relevance scores. In a simplified scenario, $\mathbb{A}$ could be a simple two-element set $\mathbb{A}$=\{\textit{Relevant, Non-Relevant}\} so that a solution maps each query-document pair to either relevant or non-relevant. In practice, $\mathbb{A}$ can be the set of positive real numbers $\mathbb{R}^+$ so as to account for real-valued relevance scores. This example shows that sometimes it is important to carefully design the question and answer sets so as to represent the solution in an appropriate way.

% ranking q -> d. However, (q,d) -> R/non-r.

Depending on the scenario, a solution can present as different forms of mappings, such as a function $S(f)$, an algorithm $S(a)$ or a model $S(m)$. A function (math equation) is the most convenient way to map questions to answers and this form is most widely used in science and engineering such as physics and mechanical engineering. But many solutions can be too complicated to be expressed as a function, especially in computer science and most notably in AI. To solve the problem, algorithms map questions to answers through a procedure, which can be considered as multiple steps of (sometimes nested) functions. This algorithmic-type of mapping is most widely used in theoretical computer science and algorithms research. In some other cases, a mapping function exists but the exact analytical form of the function is very difficult to find. In these cases, we initialize the mapping function as a model architecture such as a deep neural network due to the Universal Approximation Theorem (UAT) \cite{cybenko1989approximation,hornik1991approximation,csaji2001approximation} and then ``learn'' the parameters of the model based on observational or counterfactual examples. The final learned model thus serves as the mapping from questions to answers. This model-type of mapping is most widely used in AI/ML research. More often, though, the mapping is a combination of functions, algorithms and models due to the complex nature of many problems and their solutions.

% algorithm theoretical computer science.
% model: AI/ML.
% model can be considered as a function, but we use them separately for convenience.
% $S(f), S(a), S(m)$

Another note to make is that the definition of solution itself does not imply anything about the quality of the solution. A solution can be either good or bad or even stupid, but they all can be solutions. Seeking for good solutions depends on the definition of ``what is good'', i.e., the evaluation method, but the definition of solution does not involve evaluation as part of it. Instead, deciding how good a solution is goes to the evaluation module, which is largely independent from the solution itself because we can evaluate a solution from whatever perspective we want depending on our need. 
% Besides, this paper cares more about what is a good problem rather than what is a good solution to the problem, while seeking for good solutions to a given problem is the responsibility of existing AI methods through representations, models, losses, optimizers, etc. (Figure \ref{fig:architecture}), though sometimes the two concepts may rely on each other.
Besides, seeking for good solution to a given problem is the key focus of existing AI methods through representations, models, losses, optimizers, etc. (Figure \ref{fig:architecture}), however, this paper cares more about what is a good problem rather than what is a good solution to the problem, though sometimes the two concepts may rely on each other. 
We will further exposit this point in the following sections of the paper. 
% \yz{TODO}

% It just a mapping. any mapping is a solution.
% unsupervised, supervised.

\subsection{Problem}

Depending on whether the potential answer set $\mathbb{A}$ is pre-defined or yet to be found, a problem can be defined in either deterministic or non-deterministic manner.

\begin{tcolorbox}[title = Problem (P)]
Deterministic Problem (DP): \textit{Given} the question set $\mathbb{Q}$ \textit{and} the answer set $\mathbb{A}$, try to find the optimal solution $S^*: \mathbb{Q} \rightarrow \mathbb{A}$ that maximizes (or minimizes) an evaluation metric or protocol $\mathcal{M}$.
\tcblower
Non-Deterministic Problem (NDP): \textit{Given} the question set $\mathbb{Q}$, try to find the potential answer set $\mathbb{A}$ \textit{and} the optimal solution 
% (i.e., mapping function) 
$S^*: \mathbb{Q} \rightarrow \mathbb{A}$ that maximizes (or minimizes) an evaluation metric or protocol $\mathcal{M}$.
\end{tcolorbox}

Note that the meaning of deterministic vs. non-deterministic here is different from that in algorithm complexity theory. In the problem learning context, deterministic problems are those that the potential answers to the problem are known. For deterministic problems, the questions can be either general questions whose potential answers are \textit{Yes} or \textit{No}, or special questions that admit many possible answers, but in either case, the potential answer set is provided and the problem is only to find a good mapping between the question and answer sets. However, for non-deterministic problems, the answer set is unknown and finding the answer set is part of the problem.
% They can also be special questions 

Take image classification as an example. Under deterministic problem setting, the questions in the question set can be general questions such as ``\textit{is this image a cat}'' and in this case the answer set $\mathbb{A}$=\{\textit{Yes}, \textit{No}\}. The question set can also consist of special questions such as ``\textit{what's the class of the image}'' and in this case the answer set must be provided to qualify as a deterministic problem, such as $\mathbb{A}$=\{\textit{cat, dog, horse, $\cdots$}\}. If the answer set is unknown and finding the answer set is part of the problem, then it will be a non-deterministic problem. An intuitive but not necessarily complete analogy for better understanding is supervised learning vs. unsupervised learning under the machine learning context, where a typical deterministic problem is supervised learning such as classification and a typical non-deterministic problem is unsupervised learning such as clustering.

% In machine learning context, deterministic problems are usually supervised learning. 

% \begin{tcolorbox}
% Non-Deterministic Problem (NDP): \textit{Given} the question set $\mathbb{Q}$, try to find the potential answer set $\mathbb{A}$ \textit{and} the optimal solution (i.e., mapping function) $S^*(f): \mathbb{Q} \rightarrow \mathbb{A}$ that maximizes (or minimizes) an evaluation metric or protocol $\mathcal{M}$.
% \end{tcolorbox}

% In machine learning context, non-deterministic problems are usually unsupervised learning such as clustering. 

% $P: \exists$

\subsection{Problem Learning}

\begin{tcolorbox}[title = Problem Learning (PL)]
% Deterministic Problem Learning (DPL): Based on data or interaction with environment, find $(\mathbb{Q}, \mathbb{A}, \mathcal{M})$ triples that constitute \textit{valid} and \textit{ethical} problems.
Deterministic Problem Learning (DPL): Based on data or interaction with environment, find $(\mathbb{Q}, \mathbb{A}, \mathcal{M})$ triple(s) such that $(\mathbb{Q}, \mathbb{A}, \mathcal{M})$ constitutes a \textit{valid} and \textit{ethical} deterministic problem.
\tcblower
% Non-Deterministic Problem Learning (NDPL): Based on data or interaction with environment, find $(\mathbb{Q}, \mathcal{M})$ pairs that constitute \textit{valid} and \textit{ethical} problems.
Non-Deterministic Problem Learning (NDPL): Based on data or interaction with environment, find $(\mathbb{Q}, \mathcal{M})$ pair(s) such that $(\mathbb{Q}, \mathcal{M})$ constitutes a \textit{valid} and \textit{ethical} non-deterministic problem.
\vspace{-1ex}
\end{tcolorbox}

Basically, problem learning aims at proposing problems rather than solving problems but the proposed problems should be valid and ethical. This can be seen as a constrained learning problem where the constraint consists of validity and ethics requirements.
Though problem learning does not directly solve the proposed problem, it can be closely related to solving the problem because one important aspect of ``valid'' is whether the proposed problem is solvable at all. However, in many cases, we do not have to really solve the problem to decide if the problem is solvable at all, instead, we can adopt various methods to test the solvability of a problem before we really devote efforts to solve it. We will use separate sections to discuss in detail what do we mean by ``valid'' and ``ethical'' as well as how to find such valid and ethical problems. In this section, we focus on introducing the definition and intuition of problem learning.

% We start from deterministic problem learning (DPL). DPL aims to find $(\mathbb{Q}, \mathbb{A}, \mathcal{M})$ combinations under some problem validity and ethics constraints. 
We use a toy example to illustrate what problems we can possibly expect to be proposed by a problem learning agent. Suppose the problem learning agent has seen two problems before, one is an image classification problem $(\mathbb{Q}_1, \mathbb{A}_1, \mathcal{M})$ and the other is an emotion classification problem $(\mathbb{Q}_2, \mathbb{A}_2, \mathcal{M})$, where $\mathbb{Q}_1$ includes various images and $\mathbb{A}_1$=\{\textit{cat, dog, horse, $\cdots$}\} includes various image labels, $\mathbb{Q}_2$ includes various sentences and $\mathbb{A}_2$=\{\textit{happy, sad, fear, anger, disgust, surprise, $\cdots$}\} includes various emotion labels, and $\mathcal{M}$ is some classification accuracy metric shared by the two problems. Based on certain problem learning methods (such as problem composition) which we will discuss in the later sections, the agent may propose a new problem $(\mathbb{Q}_1, \mathbb{A}_1\times\mathbb{A}_2,\mathcal{M})$, which probably can be called as an ``emotional image classification'' problem. The problem aims to classify images into emotional objects such as a ``happy cat'' or an ``angry dog''. The most straightforward approach to solving the problem is probably a two-stage classification procedure, but the problem may also invoke other innovative methods such as emotional representation learning. Through problem generalization, another problem $(\mathbb{Q}_1\times\mathbb{Q}_2, \mathbb{A}_2,\mathcal{M})$ may be proposed, which can be called a ``captioned image emotion classification'' problem, which aims to classify a captioned image into certain emotion labels. If some other question sets $\mathbb{Q}^\prime$ are available, the agent may even generalize to a broader scope of problems such as $(\mathbb{Q}^\prime,\mathbb{A}_2,\mathcal{M})$, which aims towards emotion prediction of other different types of objects.

The above examples are simple since the new problem is a composition of existing problems and humans can easily come up with such problems. Actually, there have already been some research on different but similar problems such as image sentiment classification \cite{you2015robust,hu2018multimodal,yang2018weakly}. However, when problem learning is applied in non-trivial scenarios, we may discover new problems that are beyond imagination.
% but we hope machines can. 
One example is problem discovery from plain data---especially under self-supervised learning settings---rather than from previously seen problems as above. In this case, problem learning can tell what meaningful (valid and ethical) problems can be proposed based on the raw data collected from various sources in various domains.
% such as smart city and smart health.
% and problem composition or generalization are not the only approaches to problem learning. 
% nature science problems.
For example, the agent may find new problems of human-machine interaction
% predictable features or feature combinations 
through user behavior analysis on the Web or in cyber physical systems; detect new health problems or phenomena that are worth study by collective anomaly detection in combinations of metabolic indicators; or discover new signals that are worth predicting when interacting with the environment.

% In AI research domains such as , a problem learning agent may discovery problems such as NLP, CV, IR, Robotics, ... is worth predicting or a ... problem is worth solving, and then leverage representation / model / loss learning etc. to solve the problem.

Perhaps one of the most exciting application scenarios of problem learning is in science discovery, including both natural science research such as physics, chemistry, biology, drug discovery and medical research, as well as social science research such as economics, psychology and sociology. In these areas, raising new and meaningful problems---even without solving the problems---may greatly change researchers' views and inspire new ideas or research directions. One example is problem learning from failures---when the agent finds that using existing methods cannot predict something well, then a new problem that is worth studying may be raised.
For example, by learning from the massive data collected by Large Hadron Collider (LHC), a problem learning agent may discover problems such as the observed interaction among a combination of particles deviates from the predictions of existing theory, which could hint at possible new physics phenomena. This particular combination of particles may be difficult to manually detect from massive data or may be out of the expectation of researchers. 
% In drug discovery, 
% In smart farming scenarios, patrol farming robots may be able to detect detecting . detect and report potential problems unknown but abnormal phenomina in the field.
% might be predictable and this particular (perhaps complicated) combination has never been considered by researchers.
% economic research such as to detect what complex human behaviors are predictable.
% problem: prediction is not accurate. current methods can not predict something well. then we can come up with a problem.
% problem learning through interaction with environment. naturally arose problem. current methods can not predict something well. then we can come up with a problem.

Problem learning is also important in terms of helping to fill the community gaps. In the modern academic world, the amount of human knowledge has grown exponentially compared to that of hundreds of years ago, and thus it has been almost impossible for any researcher to possess knowledge from all disciplines. As a result, it frequently happens that one community has significantly advanced the definition of a problem or the techniques to solve the problem, but another community is still taking on outdated problem definitions or using outdated techniques to solve the problem. Problem learning agent can help to reduce the community gap by maintaining a global problem space aggregated from various disciplines. When raising problems in one community, the agent will leverage insights from other communities' problem definitions. With the help of problem learning agents, researchers in one community---when trying to identify important problems to solve in his or her home community---will not be limited to the scope of problems or techniques in his or her home community or any other single community.

% Communities use different techniques.  Sometimes, one community has advanced technique a lot but another community is still using a very old technique or studying a very old problem

The above examples only considered what possible problems may be discovered, but did not consider the validity and ethics of the discovered problems. We will talk about these two perspectives with more detail in the following section. One note to make here is that the evaluation metric or protocol $\mathcal{M}$ is an important part of problem learning, which helps to guarantee that
% we know how to evaluate a potential answer to the question (
a potential solution to the discovered problem can be verified. Making sure that a solution (no matter good or bad) to the problem can be verified is an important aspect of validity for the discovered problems because we hope the agent can
% that  problems are valid problems. 
avoid raising pseudo-problems.
% some problems are invalid, that's why we need $\mathcal{M}$, e.g., 
For example, G{\"o}del's incompleteness theorems \cite{godel1931formal} show that in any consistent formal system within which a certain amount of arithmetic can be carried out, there are statements which can neither be proved nor disproved \cite{sep-goedel-incompleteness}. Another well known example is the halting problem \cite{davis2013computability,turing1937computable,kleene1952introduction}, which is one of the first cases of decision problems proven to be unsolvable \cite{copeland2004essential,turing1954solvable}. Ideally, the set of evaluation metrics or protocols $\mathcal{M}$ would be able to serve two purposes: 1) evaluating any potential solution to the learned problem, which may be relatively easier compared with 2) determining if the leaned problem is solvable at all, which can be difficult but is essential to problem learning.

% would not only be able to to detect the  

% However, some problems can never be exactly determined to be solvable or not, such as the halting machine problem.

% The continuum hypothesis is a statement in the language of ZFC that is not provable within ZFC, so ZFC is not complete. In this case, there is no obvious candidate for a new axiom that resolves the issue.

Another useful discussion is the relationship between deterministic and non-deterministic problem learning. The main difference between them is whether the problem learning agent provides the candidate answer set for the discovered problems. This implies a possible trade-off between the difficulty of raising a problem and the difficulty of solving a problem (or asking a question vs. answering a question). Raising an (open) problem without providing the candidate answers could be easier, but solving such a problem becomes difficult because seeking the answer set becomes part of the problem solving procedure; on the other hand, raising a problem and meanwhile providing the candidate answers may be difficult, but solving the problem becomes easier. As a naive example, if the answer set is provided as \{\textit{Yes, No}\}, then even a random guess policy would have 50\% probability to correctly answer the question, though correctly answering a question does not necessarily mean the solver really understands the question. Overall, we present the following No Free Lunch conjecture for problem learning.

% then solving the problem may be easier, but raising such as problem 

\begin{tcolorbox}[title = No Free Lunch Conjecture]
Proposing a non-deterministic problem can be easier but solving a non-deterministic problem can be harder;
Proposing a deterministic problem can be harder but solving a deterministic problem can be easier. \end{tcolorbox}

\section{What is a Good Problem}

Problem learning can be seen as a constrained learning task. As mentioned above, a good problem should be both valid and ethical, which means that we need to add validity and ethics constraints to the learning task. We discuss these two perspectives in this section.

\subsection{Valid Problems}

There are many dimensions to talk about if a problem is valid or not. In this section, we mainly talk about problem validity from two dimensions: whether the problem is mathematically valid and whether the problem is socially valid.

\textbf{Mathematical Validity}

Mathematical validity mainly cares about if a problem is solvable at all. Usually, we expect problem learning agents to raise problems that are potentially solvable because we hope the agent can bring practical impacts. 
% mainly consider problem learning from a practical perspective. 
However, we do acknowledge that many unsolvable problems are also important because they may inspire new insights and new findings, especially from theoretical point of view. Mathematical validity can be described on two perspectives: 1) from the model perspective, we can consider the predictability of the target, and 2) from the problem perspective, we can consider the solvability of the problem.
% testing from model perspective: solvability testing.

The ability to make correct predictions is one of the most representative type of intelligence pursued by human beings. Many problems can be formulated as some type of prediction problems, such as predicting human or system behaviors, predicting object motions, and predicting certain property of the targeted items. Though many models are developed with the aim to make predictions as correct as possible, we should take care that some targets may not be predictable due to theoretical constraints. One prominent example is the chaos phenomenon, where small deviations in the initial state may lead to significant differences in the outcome \cite{lorenz1972predictability,ott1990controlling}. This is especially important in complex systems such as economics, human-in-the-loop AI systems (social networks, recommender systems, etc.), and cyber-physical systems (smart city, courier scheduling, etc.). 
Actually, predictability research has been one of the key topics in climate research \cite{kalnay2003atmospheric,webster1998monsoons,schneider1999conceptual}, economics \cite{grunberg1954predictability} and human mobility analysis \cite{song2010limits}.
In practice, some problems may be less sensitive to the deviations in the predicted outcomes, such as ads clicking prediction, product purchase prediction, movie watching prediction, while some other problems may be more sensitive to the deviations in the predicted outcomes, especially in high-stake applications that are related to people's health (e.g., smart health applications), safety (e.g., autonomous driving), reputation (e.g., intelligent legal assistants), and financial situations (e.g., loan application, retirement planning). As a result, if would best if the problem learning agent can identify the predictability \cite{song2010limits,bialek2001predictability} through predictability testing when raising prediction-type problems. 

Solvability testing, on the other hand, is more suitable for many theoretical problems especially those raised under logical languages. As mentioned above, G{\"o}del's incompleteness theorems show that in any consistent formal system there are statements which can neither be proved nor disproved. A valid problem needs to be solvable under solvability tests. Though general solvability testing methods may not exist \cite{turing1954solvable}, one can possibly develop approximate, domain-specific, heuristic-driven solvability testing methods for the raised problem. Sometimes, a problem may be theoretically solvable but not practically solvable due to many practical factors such as computational power, data availability and legal regulations, which need to be considered when determining the solvability of the raised problem. 

% The ability to make correct predictions for the future is one of the most essential pursuit humans 
% Many problems can be formalized as prediction problems, 
% One approach is predictability test, chaotic systems. Predict something. 
% which looks at the raised problem from the problem level. This
% valid can also be determined through mathematical perspectives, Godel.
% function analysis. is the candidate problem solvable? What is the complexity. (sometimes theoretically solvable but not practically solvable, due to computational power, legal regulations, culture, etc.)

% Predictability testing. Can we develop theories or models to test if something is predictable at all rather than blindly developing models to predict it.

% However, unsolvable problems are also very important.

\textbf{Social Validity}

Problem validity also needs to be considered in terms of social perspectives. A valid problem in one social context may not be valid in another social context.
The influence of social context may differ in terms of time, for example, planet motion prediction was once an invalid problem due to religious regulations hundreds of years ago, but later it becomes a valid and actually very important problem along the advancement of human civilization. 
The influence of social context may also differ in terms of space, for example, location tracking and prediction may be an invalid problem for normal persons living in normal conditions due to privacy concerns, however, for certain workers working in dangerous areas or under dangerous conditions, location tracking and prediction could be an extremely important problem to protect their safety. 
As a result, problem learning agent should be able to raise valid problems according to the social context in which the problem is being proposed. A far-sighted agent may even raise problems that seem to be invalid now but may become valid in the future so that we'd better consider it early.

% valid -> invalid. invalid -> valid.
% culture but not a (valid) problem in another culture. e.g.,  predict the 
% e.g., through problem generalization, machine has seen two type of data: animal ..., human ..., -> human ...? some are valid, some are not, some are ethical, some are not.
% a mixture of data, various features, 

% \vspace{-3ex}
\subsection{Ethical Problems}

Ethical considerations in problem learning is very important, because we hope machines with free will can help humans rather than hurt humans. This requires AI to seek for problems that are non-maleficence, responsible, socially beneficial and that advocate the human dignity. Ethical constraints for problem learning can be considered on several dimensions, including but not limited to transparency and explainability, fairness and justice, accountability and robustness, as well as privacy and security.
% as well as sustainability and robustness.
% fairness, justice, explainability, accountability, sustainability, robustness, trustworthiness, transparency, privacy and security.
% reliability

\textbf{Transparency and Explainability}

An ideal problem learning agent would be able to explain why a certain problem is raised and why the problem is important or worth solving. This will be very helpful for humans to understand the behavior of the agent and build trust with the intelligent machines \cite{zhang2020explainable,gunning2019xai}. Over the past years, many explainable AI methods have been developed, including model intrinsic explanation methods such as linear regression \cite{montgomery2021introduction}, decision tree \cite{breiman2017classification,quinlan1986induction} and explicit factorization \cite{zhang2014explicit}, as well as  model agnostic explanation methods such as counterfactual explanation \cite{tan2021counterfactual}, feature attribution \cite{ribeiro2016should} and Shapley values \cite{shapley201617,sundararajan2020many}. However, most of the methods discuss explainable AI on the model perspective instead of the problem perspective, i.e., they mostly focus on explaining how the decision model works, but not on why a problem is important or worth solving. As a result, these methods may be able to explain the internal mechanism of the problem generation process, but may not be able to explain why we should care about the problem. However, the later is even more important under the problem learning context, because good explanations about why a problem is important will help humans to understand the insights of the raised problem and make better decisions about what problems need to be taken care of and what problems can be ignored. Problem-level explanations can be provided on both technical perspectives and social perspectives. Technically, the agent can explain what factors triggered the generation of the problem; socially, the agent can explain what social impacts will be brought if the raised problem is solved.
% on the importance of the raise problem so as to understand 

\textbf{Fairness and Justice}

Problem learning agents should take care not to raise problems that may discriminate or unfairly treat certain individuals or certain groups of people. Fairness of AI has gained attention from researchers in the recent years. Many fairness criteria and fairness-aware algorithms have been proposed, including both group fairness \cite{pedreschi2009measuring} and individual fairness \cite{dwork2012fairness}, both associative fairness \cite{yao2017beyond} and causal fairness \cite{kusner2017counterfactual}, both static fairness \cite{li2021user} and dynamic fairness \cite{ge2021towards}, as well as both single-side fairness \cite{li2021towards} and multi-side fairness \cite{burke2018balanced}. They also have been considered in various tasks such as classification \cite{dwork2012fairness,woodworth2017learning,zemel2013learning}, ranking \cite{singh2018fairness,biega2018equity,ekstrand2021fairness}, recommendation \cite{ge2021towards,li2021towards,lin2017fairness,zhu2018fairness}, cyber-physical systems \cite{wang2020faircharge}, healthcare applications \cite{agniel2018biases}, online recruiting \cite{geyik2019fairness} and financial services \cite{lee2014fairness}. However, current research on fairness in AI are mostly conducted on the model-level or result-level, i.e., they usually focus on the fairness of the machine learning model or fairness of the model decision results, but not too much on the fairness of the problem itself. Problem-level fairness needs to consider whether a problem definition is fair. For example, a problem could be unfair if the problem description aims to bring benefits to certain groups of people but ignores other groups of people. A problem description may also be unfair or even unjust if it asks to scarify certain groups or individuals so as to benefit other groups or individuals. Solving unfair or unjust problems may bring more harms than benefits to our community. It does not mean that such ``sensitive'' problems should not be raised at all, because these problems may still be very helpful for us to gain understandings of the status quo and to seek possible solutions, however, such problems need to be raised in fair, just, caring and humanistic ways.

% which can be considered based on several perspectives. For example, 
% though the relationship between definition of fairness can be controversial. 
% Granting machines with the free will to define problems that they think are important does not mean AI should have the freedom to define and solve \textit{any} problem, but \textit{ethical} problems. This means that problem learning should be conducted under ethical principles \cite{jobin2019global,taddeo2018ai} 
% such as transparency, privacy, fairness, justice, trustworthiness, responsibility and sustainability. For example, some problems such as trying to predict one's sexual orientation, political stance or whether one will quit the job may need to be eliminated from the problem space due to ethical concerns, while some other problems such as predicting one's life expectancy may remain in the problem space since they could be helpful to human but they need to be handled with extreme care \cite{siegel2020when}.
% In general, AI should seek for problems that are non-maleficence, socially beneficial and that advocate human dignity.

\textbf{Accountability and Robustness}

Problem learning gives machine the ability to define and solve the problems that they think are important. However, problem learning agents may be vulnerable to malicious attacks or manipulations. An intelligent machine with the free will to seek problems to solve---if manipulated by bad-faith individuals or entities---could be very dangerous for humans, because the machine may be instructed to create unethical or harmful problems. As a result, accountability and robustness of problem learning is very important, and the problem learning agent should be able to handle unexpected errors, resist attacks, and produce consistent outcomes.

\textbf{Privacy and Security}

Problem learning should also take care of the user privacy and guarantee the security of the protected information. The reason is that the generated problem descriptions may reveal the private information of certain individuals or groups, especially for those problems that are related to processing user-generated data. For example, the problem ``what will be Alice's probability of cure if she take a certain treatment,'' though helpful for Alice, may unwillingly reveal Alice's privacy on health condition to third-parties. As a result, when treating private or sensitive information, problem learning agent should take care to avoid data leakage, protect user privacy and raise problems in safe and responsible ways.

% The above examples did not consider validity and ethics constraints. 
% human-related problems. face-recognition + what classification?
% example: a problem both valid and ethical, a problem valid but unethical, a problem ethical but invalid, a problem invalid and unethical.
% 1. Still classification, ranking, prediction problems, but new classification, ranking, prediction problems. e.g., beyond face classification, there could be other classification problems. Beyond purchase prediction, there could be other prediction problem. The key is to identify what is ``predictable'' from data.
% 2. Discover new problems beyond existing paradigms such as classification, prediction and ranking.
% 3. Ethical

%%%% Other thoughts

% find P s.t., valid and ethical.
% answer: Y/N. question: simple question, complicated question, combined question, questions in various domains (human, nature, science, etc.)
% does not solve the problem, but to find valid (solvable) and ethical problems.
% D data point, L label? can model classification, ranking, prediction, clustering, etc.

\section{Possible Approaches to Problem Learning}

The approaches towards problem learning can be broadly classified into two types: Differentiable Problem Learning ($\partial$PL) and Non-Differentiable (Discrete) Problem Learning (\sout{$\partial$}PL). Differentiable problem learning creates new problems through learning in a continuous space, while non-differentiable problem learning creates new problems through discrete problem analysis or optimization. In the following problem learning methods we discuss, some can be implemented in differentiable ways, some can be implemented in discrete ways, while some others can be implemented in both ways. We will make detailed discussions when appropriate.

% Four versions: $\partial$DPL, $\partial$NDPL, \sout{$\partial$}DPL, \sout{$\partial$}NDPL.

% \subsection{Differentiable Problem Learning ($\partial$PL)}

\textbf{Problem Learning from Failure}

In current AI research paradigms, researchers usually define a problem and then develop various models trying to solve the problem. If existing models can not solve the problem well, researchers are usually inclined to assume that existing models are not good enough and thus they devote more efforts trying to design better models. However, if existing models cannot solve the problem well, maybe it is not because the models are not good enough, but because the problem itself is not well defined, i.e., the problem may not be asked in the right way. As a result, instead of spending efforts trying to design better models, it is probably more important to think about how to define the problem in a better way. Actually, the ability to refine problem and methods iteratively is a fundamental skill in many research areas especially natural science research, and this skill needs to be captured by intelligent machines. Problem learning from failure aims at this goal. When the agent can not produce satisfactory results such as prediction accuracy on a given problem based on existing models, it would be very inspiring if the agent can propose possible modifications to the problem definition, such as ``if the problem is (slightly) changed to this new definition, then existing models will do very well on the new problem.'' For example, the agent may be originally asked to predict the preference of online users, but existing models are unable to make satisfactory predictions, however, the agent may discover that the preference of a certain sub-group of users is very predictable, and thus suggests to solve this new problem instead. 

Technically, problem learning from failure requires the ability to explore alternatives to the ``anchor problem,'' which can be achieved by either differentiable or non-differentiable learning. Usually, the anchor problem may include both numerical and categorical descriptions. For example, for the anchor problem ``rank the \textit{sci-fi} movies whose ratings are \textit{greater than 3} for all \textit{male} users for personalized recommendation,'' both categorical features \textit{sci-fi} or \textit{male} and numerical features \textit{greater than 3} can be changed to alternatives. The problem learning agent will optimize in the search space created by the alternative values of these features to learn for the best problem that leads to good recommendation accuracy based on, e.g., reinforcement learning. In this procedure, optimizing over categorical features needs discrete learning, while optimizing over numerical features needs differentiable learning. Eventually, the agent may find that the new problem ``rank the \textit{cartoon} movies whose ratings are \textit{greater than 4} for all \textit{children} users for personalized recommendation'' is a better solvable problem.

% In this context, differentiable learning mainly helps to revise the anchor problem by learning better numerical numbers 
% This later possibility is a very fundamental realizing that a problem may not be well-defined is a very important ... for natural science research, but it's not well practiced in AI research.

% \subsection{Non-Differentiable Problem Learning (\sout{$\partial$}PL)}

\textbf{Problem Learning from Exploration}

Problem learning from failure produces new problems based on existing anchor problems. Another approach to problem learning is learning from exploration. This approach does not depend on a given anchor problem, instead, the agent aims to discover valid and ethical problems by active explorations, such as exploring a dataset or interacting with the environment. One example is to actively discover predictable or approximately predictable signals from data that are previously ignored. Exploring the predictability of single-feature signals is a starting point but that may look trivial since many single-feature signals have already been manually explored by domain experts. However, it is non-trivial to explore the predictability of certain combinations of singles. It would be possible that each signal individually is not predictable, but when aggregated in certain ways, the combined signal is predictable. For example, the behavior of a single user may be very unpredictable, but the aggregated behavior of a crowd may be better predictable. Such aggregated predictors that are still unknown to researchers may exist in many areas such as social science, health and medical research, biology, environmental science, etc. If a previously unknown predictor can be identified through the problem learning agent's exploration in data or interaction with the environment, then a valid and ethical problem may be proposed for further studies.

% \textbf{Problem Learning from Interaction}

% \textbf{Problem Composition and Decomposition}

\textbf{Problem Learning by Composition}

Problem composition aims to build bigger, more ambitious problems which is composed of a sequence of smaller, well-defined and known problems. Each smaller problem in the sequence, if solved, will provide some useful information that enables the next problem to be solved. Problem composition is not as simple as putting a few smaller problem together, but needs to carefully consider the relationship between the small problems and how do they influence each other. A good problem learning agent would be able to identify the relationships between problems and connect them in the right way so that it can reach a valid problem. Problem composition can be considered as a reverse process of planning. In planning, the goal problem is given and the agent needs to decompose the problem into several smaller, easily solvable ones so as to reach the goal. In problem composition, however, the agent is provided with several small solvable problems and it aims to propose bigger problems that are valid. Such bigger problems, if solved, may provide new insights that are more than what the known small problems can provide together. In this process, the agent is also autonomously deciding what goal to reach, because different compositions of small problems may reach different goals.

% given, problem composition, AI decides what goal to reach)

\textbf{Problem Learning by Generalization}

Many known problems can be generalized to new problems. Problem generalization starts from a known problem and generalizes the problem into new ones by exploring alternatives of the subject, predicate or object in the problem description. For example, if the agent already knows that human face classification is a valid problem, it may generalize to other problems by exploring alternatives of the subject and result in new problems such as dog face or cat face classification; it may also explore alternatives of the predicate and result in new problems such as human face detection, restoration or beautification. Another example is that if the agent knows that consumer purchase prediction is a valid problem in e-commerce, then it may generalize to consumer return or consumer complaint prediction problems and conduct predictability testing to decide the validity of the problems.
% 1. Still classification, ranking, prediction problems, but new classification, ranking, prediction problems. e.g., beyond face classification, there could be other classification problems. Beyond purchase prediction, there could be other prediction problem. The key is to identify what is ``predictable'' from data.
% Problem learning by generalization is different from problem learning from failure in terms of (1) problem learning from failure does not change the target of the problem but only changes what to do with the object, while problem learning by generalization explores alternatives of the object. Supervised problem learning vs. unsupervised problem learning. 
Problem learning by generalization is connected with problem learning from failure in that once the new problem has been generated by generalization, it may or may not be well solved with carefully designed models. If the problem cannot be well solved, then one many use problem learning from failure to further refine the problem definition.

\textbf{Problem Search}

The above approaches assume that a problem is described in plain language. However, a problem can be represented using mathematical structures such as graphs. 
% There could be two approaches to representing problems using graph structures: concept-level representation and question-level representation. 
% Concept-level representation 
For example, we can take the various concepts such as ``human,'' ``face,'' ``classification,'' ``prediction,'' ``cat,'' ``dog'' and ``consumer'' as the potential entities in the graph and use edge connections between the entities to describe a problem. Usually, such entities come from the question set $\mathbb{Q}$. Once problems can be represented as graphs, we can develop Problem Architecture Search (PAS) algorithms to search for valid and ethical problems, which is similar to Neural Architecture Search (NAS) for searching good model architectures. PAS can be implemented based on reinforcement learning and the reward signals can be provided by checking the validity and the ethics of the generated candidate problems. To make problem search more controllable, we may control the set of concepts used for problem search so that the agent can generate problems within the target set of concepts. 

Problem search may also be conducted beyond concept-level. For example, problem learning by composition, as noted above, can be implemented through problem search. Specifically, each unit-problem can be considered as an entity in the graph and PAS can be adopted to search for valid and ethical compositions of the unit-problems so as to construct bigger, more ambitious problems. This can be considered as a modularized architecture search procedure.

% we can consider Modularization, assemble modularized components.

% Question-level representation 

% One is to represent the  syntactic analysis 

\textbf{Meta Problem Learning}

% The idea of meta-learning 

Though there are an infinite scope of problems and the problem definition varies from one to another, there could still exist some similarities and common structures shared by many problems. Meta Problem Learning can help to learn such common structures as ``meta-problems,'' which can then be used to induce new problems. Meta problem learning can be beneficial in terms of several perspectives. First, it can help to extract the similarities from seemingly different problems so as to enable collaborative learning effects for discovering good problem structures. Under supervised problem learning setting, i.e, when a set of known valid and ethical problems are provided as supervision, such collaborative learning effect can help the agent to learn the latent definitions of ``validity'' and ``ethics'' from the supervisions and encode them into the meta-problem, so that specific problems induced from the meta-problem can easily satisfy the validity and ethics requirements. Second, by learning meta-problems from various different domains and then generating specific problem from the learned meta-problems, it helps to enable cross-domain problem learning or problem transfer learning. Finally, by learning domain-independent meta-problem structures, it helps to improve the efficiency of problem architecture search by fine-tuning from the meta-problems instead of conducting problem search from scratch.

\bibliographystyle{unsrt}  
\bibliography{references}  %%% Remove comment to use the external .bib file (using bibtex).


% George Kour and Raid Saabne.
% \newblock Real-time segmentation of on-line handwritten arabic script.
% \newblock In {\em Frontiers in Handwriting Recognition (ICFHR), 2014 14th
%   International Conference on}, pages 417--422. IEEE, 2014.

% George Kour and Raid Saabne.
% \newblock Fast classification of handwritten on-line arabic characters.
% \newblock In {\em Soft Computing and Pattern Recognition (SoCPaR), 2014 6th
%   International Conference of}, pages 312--318. IEEE, 2014.

% \bibitem{hadash2018estimate}
% Guy Hadash, Einat Kermany, Boaz Carmeli, Ofer Lavi, George Kour, and Alon
% \newblock Estimate and replace: A novel approach to integrating deep neural
%   networks with existing applications.
% \newblock {\em arXiv preprint arXiv:1804.09028}, 2018.

% \end{thebibliography}

\end{document}